\documentclass[letterpaper, 10 pt, conference]{ieeeconf}  % Comment this line out

\IEEEoverridecommandlockouts                              % This command is only
\usepackage[colorlinks,linkcolor=green,citecolor=red,urlcolor=blue,bookmarks=true,hypertexnames=true]{hyperref} 
\usepackage[utf8]{inputenc}
\usepackage[T1]{fontenc}
\usepackage{balance}

\usepackage{graphicx} % for pdf, bitmapped graphics files
\usepackage{epsfig} % for postscript graphics files
\usepackage{mathptmx} % assumes new font selection scheme installed
\usepackage{mathptmx} % assumes new font selection scheme installed
\usepackage{amsmath} % assumes amsmath package installed
\usepackage{amssymb}  % assumes amsmath package installed
\usepackage{booktabs}
\usepackage{url}
\usepackage{multirow}
\usepackage{multicol}
\title{\LARGE \bf
Video-based Traffic Light Recognition by Rockchip RV1126 for Autonomous Driving}

\author{Miao Fan$^{1,*}$, Xuxu Kong$^{2}$, Shengtong Xu$^{3}$, Haoyi Xiong$^{4}$, and Xiangzeng Liu$^{5}$ % <-this % stops a space
\thanks{$^{1}$Chief scientist at NavInfo Co. Ltd., China. Senior member of IEEE.}%
\thanks{$^{2}$Senior engineer at NavInfo Co. Ltd., China.}
\thanks{$^{3}$Principal product manager at Autohome Inc., China.}
\thanks{$^{4}$Principal scientist at Baidu Inc., China. Senior member of IEEE.}
\thanks{$^{5}$Associate professor at Xidian University, China.}%
\thanks{*Correspondence: {miao.fan@ieee.org}}
}
        
\begin{document}

\maketitle
\thispagestyle{empty}
\pagestyle{empty}

%%%%%%%%%%%%%%%%%%%%%%%%%%%%%%%%%%%%%%%%%%%%%%%%%%%%%%%%%%%%%%%%%%%%%%%%%%%%%%%%
\begin{abstract}
Real-time traffic light recognition is fundamental for autonomous driving safety and navigation in urban environments. While existing approaches rely on single-frame analysis from onboard cameras, they struggle with complex scenarios involving occlusions and adverse lighting conditions. We present \textit{ViTLR}, a novel video-based end-to-end neural network that processes multiple consecutive frames to achieve robust traffic light detection and state classification. The architecture leverages a transformer-like design with convolutional self-attention modules, which is optimized specifically for deployment on the Rockchip RV1126 embedded platform. Extensive evaluations on two real-world datasets demonstrate that \textit{ViTLR} achieves state-of-the-art performance while maintaining real-time processing capabilities (>25 FPS) on RV1126's NPU. The system shows superior robustness across temporal stability, varying target distances, and challenging environmental conditions compared to existing single-frame approaches. We have successfully integrated \textit{ViTLR} into an ego-lane traffic light recognition system using HD maps for autonomous driving applications. The complete implementation, including source code and datasets, is made publicly available to facilitate further research in this domain.
\end{abstract}
{\keywords traffic light, real-time recognition, ego-camera videos, Rockchip RV1126, autonomous driving.}

%%%%%%%%%%%%%%%%%%%%%%%%%%%%%%%%%%%%%%%%%%%%%%%%%%%%%%%%%%%%%%%%%%%%%%%%%%%%%%%%

\section{Introduction}
\begin{figure}
\centering
  \includegraphics[width=\columnwidth]{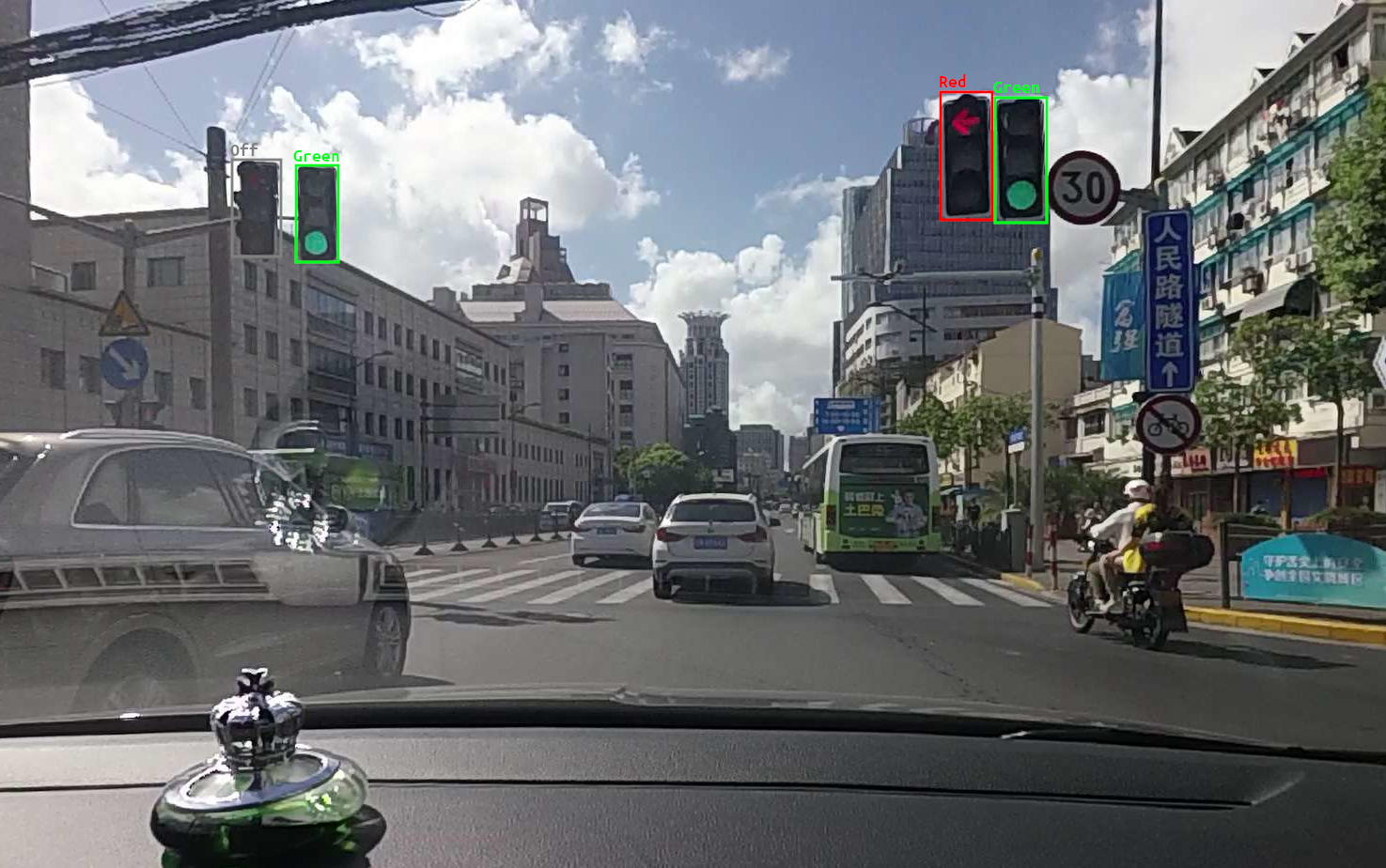}
      \vspace{-5mm}
  \caption{{A real-world example of traffic light recognition (TLR) task, including the subtasks of traffic light detection which aims at locating all traffic lights as well as classification of their states (including green, red, off, etc.). However, it is difficult to figure out which traffic light controls the ego vehicle lane in this case.}}
  \label{fig:1}
    \vspace{-5mm}
\end{figure}

Real-time traffic light recognition~\cite{7398055} based on vehicle-mounted chips plays a pivotal role in autonomous driving~\cite{5940562}, as it is a fundamental technology that empowers vehicles to locate and identify the states (red, yellow, green, etc.) of all traffic lights in a scene when crossing the intersections in urban areas. Therefore, traffic light recognition (TLR) is a compound task illustrated by Figure~\ref{fig:1}, including both detection~\cite{8569575} and state classification~\cite{yildiz2022deep} of traffic lights. As cameras become cheap alternatives to LiDARs~\cite{roriz2021automotive} for vehicle equipment manufacturers, vision-centric perception is widely regarded as a cost-effective solution to autonomous driving. Accordingly, the mainstream approaches on TLR employ the sensing images from onboard cameras~\cite{r00} and generally deploy a one/two-stage pipeline on a vehicle to perceive traffic lights in real-time. The one-stage methods~\cite{7535408,8569575,8697819,8851927} generally predict the location and state of traffic lights simultaneously via leveraging an end-to-end neural network, such as AlexNet~\cite{NIPS2012_c399862d}, SSD~\cite{liu2016ssd}, R-CNN~\cite{girshick2015region}, YOLO~\cite{7780460} and DETR~\cite{carion2020end}. On the other hand, a two-stage system~\cite{7995785,7989163,8611202} first performs finding the location of all traffic lights and then checking the signal states of the detected traffic lights. 
However, these approaches only leverage single-frame inputs and fail to exploit temporal information. As a result, they are prone to errors caused by complex environmental conditions such as occlusions, large intersections, and extreme camera exposures, which are quite frequent in autonomous driving scenarios. 

\begin{figure*}[!htp]
\centering
  \includegraphics[width=0.95\textwidth]{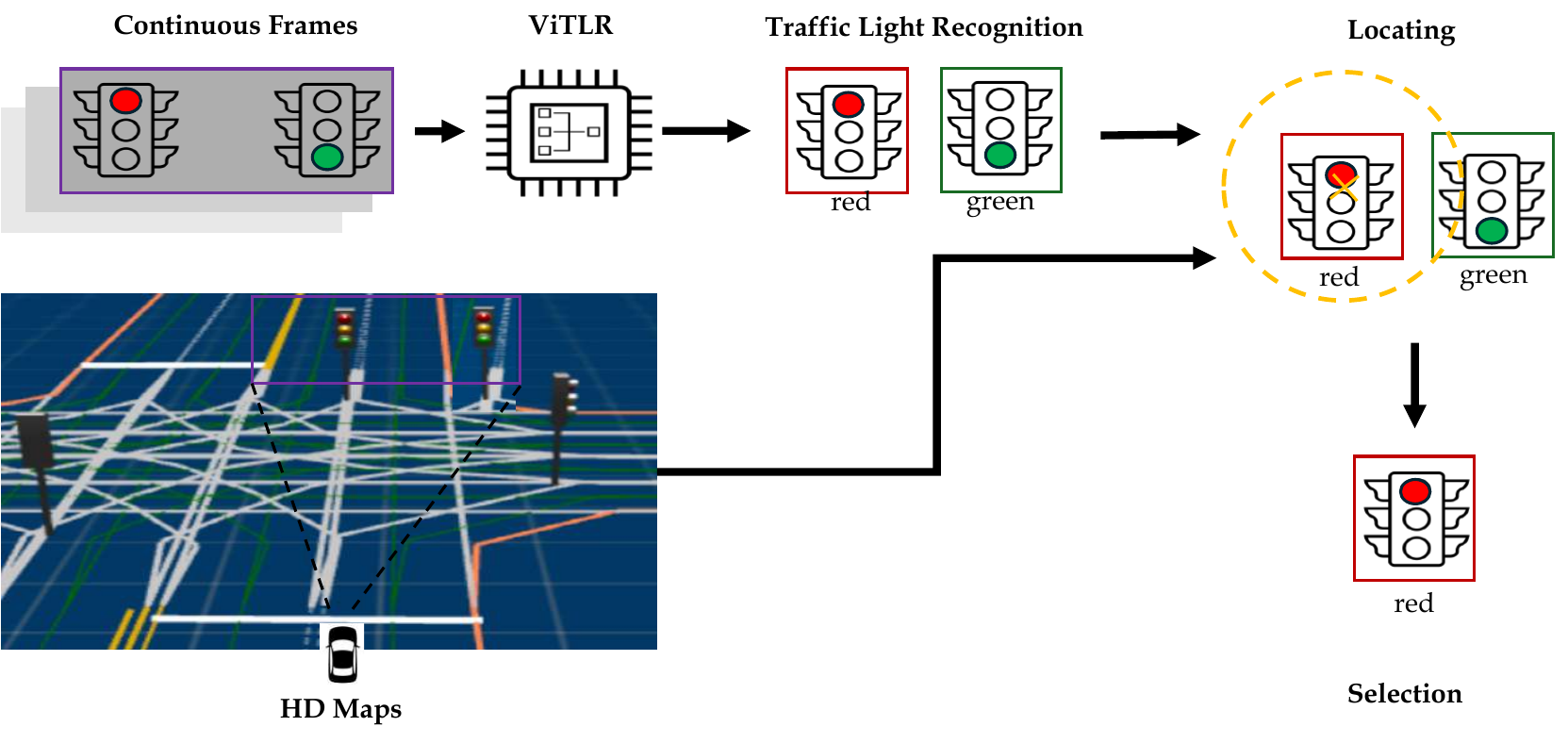}
  \vspace{-3mm}
  \caption{Overall flow of our embedded system deployed on Rockchip RV1126 SoC (system-on-a-chip) for recognizing the traffic light to ego lane with the aid of HD maps for autonomous driving. Firstly, the video stream captured by onboard cameras is continuously fed into \textit{ViTLR}, the video-based and end-to-end neural model proposed by this paper. It locates and identifies the states (red, yellow, green, etc.) of all traffic lights in real-time. Subsequently, the world position of the traffic light to the ego lane is acquired from HD maps and projected into the current frame (orange cross) based on the pose and present location of the ego vehicle. The orange circumference represents a threshold that accounts for the imprecision in localization. Finally, the closest one is selected from the bounding boxes that have centers within the threshold, and considered as final state prediction for that frame.}
    \vspace{-5mm}
  \label{fig:2}
\end{figure*}
To address the issues, we present \textit{ViTLR}, a video-based and end-to-end neural network that takes multiple frames from ego cameras as inputs, and directly outputs bounding boxes with traffic lights' states at the current frame. We devise a transformer-like~\cite{DBLP:conf/iclr/DosovitskiyB0WZ21} architecture mainly composed of convolutional self-attention (CSA) modules to drive more efficient implementation of \textit{ViTLR} deployed on Rockchip RV1126\footnote{\url{https://www.rock-chips.com/a/en/products/RV11_Series/2020/0427/1076.html}}'s NPU (neural processing unit). To the best of our knowledge, \textit{ViTLR} is the first embedded neural network deployed on a vehicle-mounted chip for real-time traffic light recognition. Since it is not capable of attributing any special meaning to the traffic lights that are relevant to the ego vehicle’s route, we need to integrate it into our ego-lane traffic light recognition system illustrated by Figure~\ref{fig:2}, with the aid of HD maps~\cite{liu2020high,r0} for autonomous driving in urban areas.

We use two real-world datasets collected from Germany and Mainland China to evaluate the performance of \textit{ViTLR} as well as several mainstream approaches on real-time TLR. Experimental results show that \textit{ViTLR} achieves state-of-the-art effectiveness under comparable efficiency on Rockchip RV1126's NPU. It also demonstrates more robust performance than the mainstream methods in further discussion of temporal stability, target distances, and complex scenarios. In order to prosper follow-up studies, we decide to release the source codes and real-world datasets to the public at OneDrive\footnote{ \url{https://1drv.ms/f/c/76645c25a8914a0b/ErYQjUadPVVNmtTb9iH2gFEB7aAyzxyeE-Q61qOMD5AgjA}} for research purposes.

\section{Related Work}
From perspectives of effectiveness and efficiency, this paper presents two contributions we made, i.e., a video-based paradigm for \textit{traffic light recognition} and its real-time neural architecture specific to \textit{Rockchip RV1126}'s NPU, to the research community of autonomous driving. Therefore, brief reviews of the latest work on {traffic light recognition} and {Rockchip RV1126} are essential. 

\subsection{Traffic Light Recognition}
The mainstream approaches on TLR can be categorized into one-stage methods~\cite{7535408,8569575,8697819,8851927} and two-stage systems~\cite{7995785,7989163,8611202}. For one-stage methods, an end-to-end neural network for object detection is commonly adopted to predict the location and state of traffic lights simultaneously. Consequently, the performance of TLR greatly depends on the specific design of neural nets for the task of object detection, such as R-CNN~\cite{girshick2015region}, YOLO~\cite{7780460} and DETR~\cite{carion2020end}, in which DETR is widely known as the SOTA approach.
The two-stage systems consist of two separate modules, where the first module identifies the location of traffic lights and the second predicts their state. 

The two-stage pipeline is considered a better alternative to implement TLR systems because the two modules can be optimized separately~\cite{8569575}. However, it leads to a bottleneck of efficiency as the two modules are sequentially stacked together. Given the computing power constraints and real-time requirements on the vehicle-mounted chip, we decided to devise a one-stage method referring to the transformer-like architecture similar to DETR for the task of real-time TLR for autonomous driving.

\subsection{Rockchip RV1126}
Rockchip RV1126 is a high-performance but low-cost vision processor SoC, especially for AI-related applications. Its built-in NPU supports INT8/INT16 hybrid operation and computing power is up to 2.0 TOPS (tera operations per second). The video encoder embedded in RV1126 supports both UHD H.265/H.264 and multi-stream encoding. With the help of this feature, a video captured from cameras can be encoded with higher resolution and stored in local memory at the same time.

The design of Rockchip RV1126's NPU adapts to the RKNN Toolkit\footnote{\url{https://github.com/rockchip-linux/rknn-toolkit}}, which does not yet support all neural operators in the major frameworks such as PyTorch and TensorFlow. When deployed, it is necessary to convert a self-developed neural network on these major frameworks into a usable RKNN model. Generally speaking, an RKNN model only supports a limited number of operators such as Dense, Flatten, Reshape, BatchNormalization, and Conv2D. Moreover, the kernel size of the Conv2D is preferably set by $3 \times 3$, for the purpose of reducing additional computational consumption of NPU and achieving maximum utilization of the chip platform.

\subsection{Video Object Detection}
Despite the maturity of object detection methods for single images, they still encounter numerous challenges in practical engineering applications. These challenges include occlusion, motion blur, improper shooting focal length, non-rigid deformation, and so on. In the context of autonomous driving, it is crucial to explore how to utilize the image information from historical frames in video streams to correct the object detection results of the current frame.

The existing video object detection method, YOLOV++~\cite{shi2024yolovpp}, generates multi-frame results through a two-stage training process. However, it focuses on optimizing the results across all frames rather than specifically processing the results of the current frame. Autonomous driving tasks, which demand real-time responses, rely heavily on the accuracy of the current frame's detection results. Moreover, some advanced structures, such as DiffusionVID~\cite{diffusionvid}, TransVOD~\cite{he2021end,zhou2022transvod}, and VisTR~\cite{wang2020end}, which are based on the Transformer architecture, are too complex to be deployed on edge devices with limited computing power. In contrast, our proposed method leverages images from historical frames to refine the detection results of the current frame, thereby reducing computational complexity and simplifying the task. 

In the current research, ConvNeXt~\cite{liu2022convnet} has demonstrated that stacking depthwise separable convolutions can achieve performance comparable to that of Transformers. The study of CvT~\cite{Wu2021CvTIC} has shown that the projection using convolution can replace multi-head self-attention for modeling spatial context.

\section{Formulation}
\begin{figure*}[htp!]
\centering
  \includegraphics[width=0.98\textwidth]{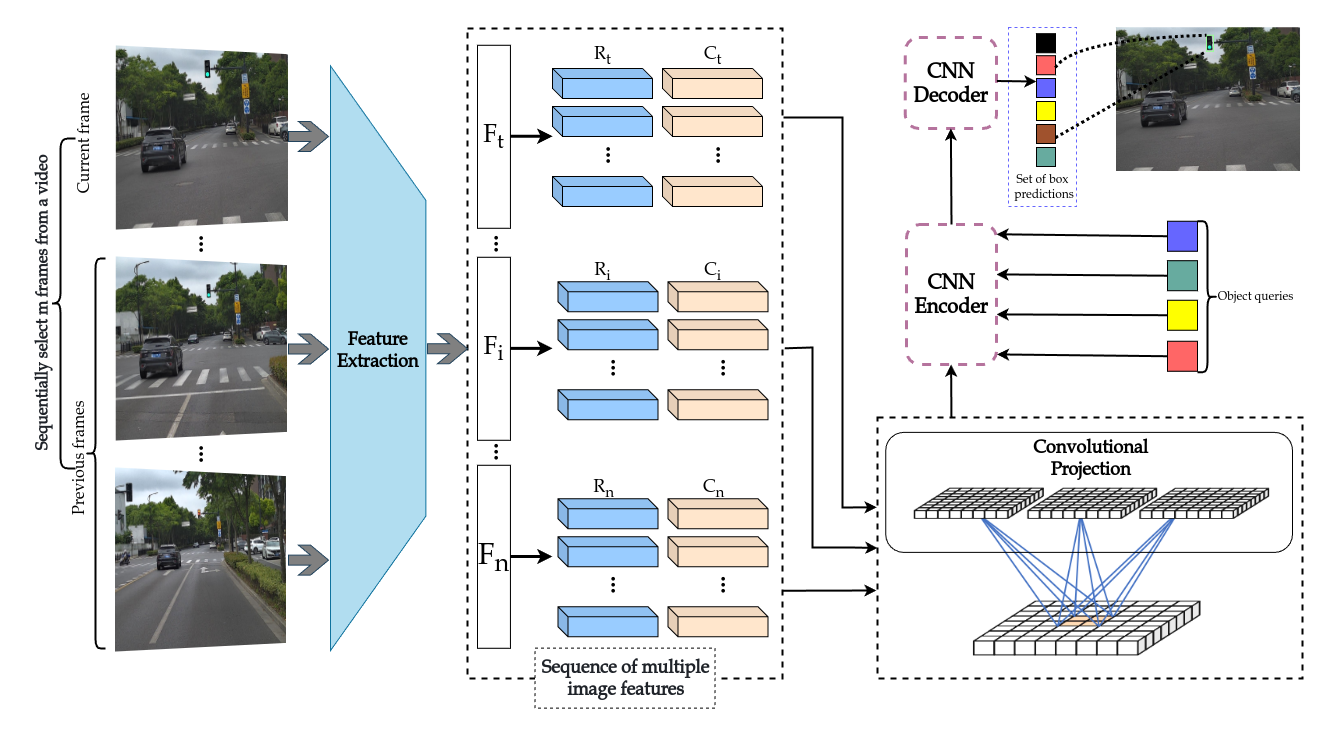}
   \vspace{-3mm}
  \caption{The architecture and training procedure of \textit{ViTLR}. We propose a transformer-like architecture mainly composed of convolutional self-attention neural modules to model \textit{ViTLR}, which can take consecutive frames captured by ego-vehicle cameras as inputs and directly output bounding boxes as well as states of traffic lights running on Rockchip RV1126's NPU within the latency of 40 milliseconds.}
    \vspace{-5mm}
  \label{fig_3}
\end{figure*}

As \textit{ViTLR} is a one-stage and end-to-end neural architecture, we use $\mathcal{F}$ to denote itself as a function and use $\Theta$ to represent its parameters.
The basic function of $\mathcal{F}(\Theta)$ is to establish a mapping from camera-captured frames as inputs to recognized traffic lights as outputs at the same timestamp $t$. If we define $\mathcal{X}_t$ and $\mathcal{Y}_t$ as inputs and outputs, respectively, then \textit{ViTLR} can be formulated by Eq.~\ref{eq1} as follows,
\begin{equation}
\label{eq1}
\mathcal{Y}_t = \mathcal{F}(\mathcal{X}_t; \Theta).
\end{equation}

One feature of \textit{ViTLR} is taking multiple frames, including the current and the most recent $n$ frames, captured by ego cameras as inputs. Therefore, $\mathcal{X}_t$ should be defined by Eq.~\ref{eq2} as follows,
\begin{equation}
\label{eq2}
\mathcal{X}_t = (\textbf{X}_{t},~...,~\textbf{X}_{i},~...,~\textbf{X}_n),
\end{equation}
where $\textbf{X}_t \in \mathbb{R}^{h \times w \times c}$ stands for a 
$c$-channels image with $h$ pixels in height and $w$ pixels in width captured at timestamp $t$.

Another feature of \textit{ViTLR} is to simultaneously output the bounding boxes as well as the states of traffic lights in the current frame. Supposing that the maximum number of traffic lights in a frame at timestamp $t$ is set by $m$, the locations of all the bounding boxes should be denoted by a tuple, which contains the center coordinates $\textbf{P}_t \in \mathbb{R}^{m \times 2}$, the height $\textbf{h}_t \in \mathbb{R}^m$ and the width $\textbf{w}_t \in \mathbb{R}^m$ in pixels of all the bounding boxes, and the predicted states of each bounding box can be defined by $ \textbf{S}_t \in \mathbb{R}^{m \times l}$, where $l$ represents the number of states.
To the end, Eq.~\ref{eq3} shows the formulation of the output $\mathcal{Y}_t$ at timestamp $t$ as follows,
\begin{equation}
\label{eq3}
\mathcal{Y}_t = (\textbf{P}_t, \textbf{h}_t, \textbf{w}_t, \textbf{S}_t).
\end{equation}

\section{Architecture}
As Rockchip RV1126’s NPU is dedicated to highly optimized convolutional operations, we devise a transformer-like architecture mainly composed of convolutional self-attention modules to implement $\mathcal{F}(\Theta)$ (i.e., the model of \textit{ViTLR}) to fulfill the real-time requirement of \textit{ViTLR} running on Rockchip RV1126.
Figure~\ref{fig_3} illustrates the architecture of $\mathcal{F}$ as well as the training procedure to acquire all the parameters (i.e., $\Theta$) of $\mathcal{F}$.

\subsection{Stream Input}
In the context of traffic scene object detection, the primary data modality is video. However, the mainstream object detection models currently in use are designed to perform localization and classification on a single frame of an image. While this approach has achieved satisfactory results in many aspects, it still encounters numerous challenges in complex real-world traffic scenarios. These challenges necessitate leveraging the relationships between multiple frames to achieve more accurate detection outcomes. To this end, we have developed a system that supports the streaming input of multiple images, enabling continuous input of video data, including both the current frame and previous frames as indicated in Eq.~\ref{eq2}.

\subsection{Neural Model}
After receiving the input of multiple frames, the feature extraction module utilizes the backbone architecture identical to that of the YOLOX model to perform feature extraction. This procedure generates three distinct sets of features: the current frame $F_{t}$, any arbitrary frame $F_{i}$, and the final frame $F_{n}$. Each set encompasses both regression and classification features. The sequence of multiple image features $F$ can be represented as Eq.~\ref{eq4}.

\begin{equation}
\label{eq4}
F=\sum_{k=t}^{n} {F}_k = \sum_{k=t}^{n} \{{R}_k,{C}_k\}
\end{equation}

Following the extraction of localization and classification features from multiple frames, we employ a Convolutional Projection~\cite{Wu2021CvTIC} module to establish spatial relationships across frames. Notably, the Convolutional Projection module leverages depthwise separable convolution in place of the conventional multi-head attention mechanism. This substitution transforms the linear projection of original positions into a convolutional projection, thereby enhancing computational efficiency while maintaining robust performance. During the computation within the Convolutional Projection module, the extracted features $F$ are initially reshaped into a 2D token map. A depthwise separable convolution with a kernel size of $s \times s$ is then applied to perform the Convolutional Projection. Finally, the projected tokens are flattened into 1D for subsequent processes. This procedure is formally expressed as follows:

\begin{equation}
\label{eq5}
E_i = \text{Flatten}(\text{Conv2d}(\text{Reshape2D}(f_i), s))
\end{equation}

where $E_i$ is the input For CNN-Encoder module, Conv2d is a depthwise separable convolution~\cite{Xception} implemented by: Depth-wise Conv2d $\rightarrow$  BatchNorm2d $\rightarrow$  Point-wise Conv2d, and $s$ refers to the convolution kernel size.

While target detection models based on the Transformer architecture have demonstrated significant performance, they face substantial challenges when deployed on low-computation edge devices. In this study, we introduce a novel purely convolutional network architecture that closely mirrors the structure of the Transformer. This architecture comprises two key components: a CNN-Encoder and a CNN-Decoder~\cite{chen2023deco}. 

We use the ConvNeXt~\cite{liu2022convnet} blocks to build CNN-Encoder. Specifically, each encoder layer is stacked with a $7\times7$ depth-wise convolution, a LayerNorm layer, a $1\times1$ convolution, a ReLU activation, and another $1\times1$ convolution.

\begin{figure}[htp!]
  \includegraphics[width=0.98\columnwidth]{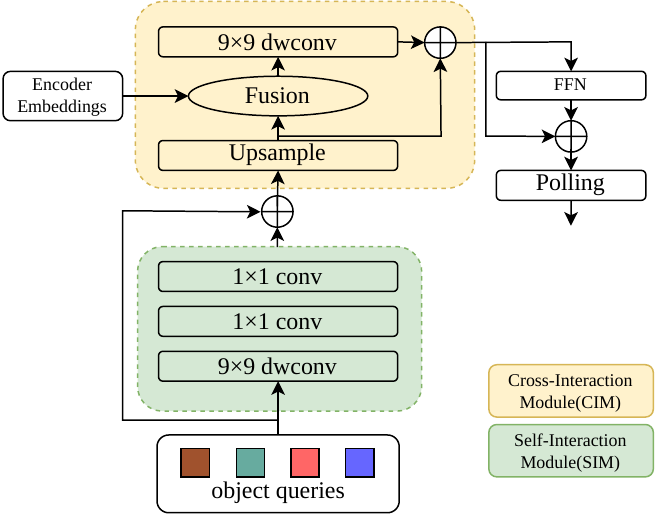}
  \caption{The CNN-Decoder~\cite{chen2023deco} adopted by ViTLR.}
  \vspace{-5mm}
    \label{fig_4}
\end{figure}

The CNN Decoder is composed of the Self-Interaction Module (SIM) and the Cross-Interaction Module (CIM),  which makes final detection prediction via a feed-forward network (FFN), as shown in Figure~\ref{fig_4}. The SIM is responsible for the interaction among object queries, while the CIM handles the interaction between object queries and image features. The inputs to the CNN Decoder are the outputs from the CNN Encoder, and the outputs are the processed object query features, which are subsequently fed into the detection head. This process is expressed by Eq.~\ref{eq6} as follows:

\begin{equation}
\label{eq6}
Prediction=\text{Polling}(\text{FFN}(\text{CIM}(\text{SIM}(F))))
\end{equation}

By encoding and decoding feature sequences, the proposed architecture effectively enables the localization and classification of targets within the current frame.

\begin{table*}[htp!]
\vspace{3mm}
\label{tab:1}
\caption{Statistics of two real-world datasets to assess the performance of TLR methods.}
  \vspace{-3mm}
\begin{center}
\begin{tabular}{l|rrr|rrr}
\toprule
\multirow{2}{*}{\textbf{Item}} & \multicolumn{3}{c|}{\textbf{DTLD}~\cite{8460737}}                                        & \multicolumn{3}{c}{\textbf{NTLD}}                                        \\ 
                      & \multicolumn{1}{c|}{\textit{Train}} & \multicolumn{1}{c|}{\textit{Valid}} & \multicolumn{1}{c|}{\textit{Test}} & \multicolumn{1}{c|}{\textit{Train}} & \multicolumn{1}{c|}{\textit{Valid}} & \multicolumn{1}{c}{\textit{Test}} \\ 

                      \midrule
\# (Video Clips)                 & \multicolumn{1}{r}{1478} & \multicolumn{1}{r}{211} & \multicolumn{1}{r|}{632}& \multicolumn{1}{r}{2500} & \multicolumn{1}{r}{576} & \multicolumn{1}{r}{895}   \\  
\# (Frames)                      & \multicolumn{1}{r}{28,525} & \multicolumn{1}{r}{4075} & \multicolumn{1}{r|}{12,453}& \multicolumn{1}{r}{58,462} & \multicolumn{1}{r}{16,543} & \multicolumn{1}{r}{29,874}   \\ 
\# (Traffic Lights)                   & \multicolumn{1}{r}{200,263} & \multicolumn{1}{r}{28,609} & \multicolumn{1}{r|}{91,982}& \multicolumn{1}{r}{512,642} & \multicolumn{1}{r}{125,632} & \multicolumn{1}{r}{235,975}   \\ 
\# (Traffic Light States)                      & \multicolumn{1}{r}{6} & \multicolumn{1}{r}{6} & \multicolumn{1}{r|}{6}& \multicolumn{1}{r}{4} & \multicolumn{1}{r}{4} & \multicolumn{1}{r}{4}   \\ 
\bottomrule
\end{tabular}
\end{center}
  \vspace{-5mm}
\end{table*}

\subsection{Loss Function}
The loss function (i.e., Eq.~\ref{eq7}) of \textit{ViTLR} consists mainly of two components: classification loss and localization loss. Given the significant imbalance in the sample distribution within the traffic light dataset, where yellow lights and off lights infrequently occur in real-world traffic scenarios, the classification task uses focal loss (Eq.~\ref{eq8}) to mitigate this issue. For the location of the bounding box, this study utilizes the CIoU loss (Eq.~\ref{eq9}), which comprehensively considers the overlap area, the center point distance, and the aspect ratio of the detection boxes, allowing for a more precise measurement of the similarity between the predicted and ground-truth boxes.
\begin{equation}
\label{eq7}
\mathcal{L} = \mathcal{L}_{focal} + \lambda \mathcal{L}_{CIoU}
\end{equation}
where \(\lambda\) is the balancing parameter.
\begin{equation}
\label{eq8}
\mathcal{L}_{focal} = -\sum_{i} \alpha_t (1 - p_t)^\gamma y_i \log(p_t)
\end{equation}
where \( y_i \) is the ground-truth label (0 or 1). \( p_t \) is the predicted probability. \( \alpha_t \) is the balancing weight, used to adjust the weights of positive and negative samples. \( \gamma \) is the modulating factor, used to control the weights of easy and hard samples.

\begin{equation}
\label{eq9}
\mathcal{L}_{CIoU} = \text{IoU} - \frac{d^2}{c^2} - \alpha v
\end{equation}
where \(\text{IoU}\) denotes the Intersection over Union between the predicted and ground-truth bounding boxes, \(d\) represents the distance between the centers of the predicted and ground-truth bounding boxes, \(c\) is the diagonal length of the smallest enclosing rectangle that covers both bounding boxes, \(v\) is the penalty term for the aspect ratio, and \(\alpha\) is the balancing coefficient.
\begin{table*}[htp!]
\label{tab:2}
\caption{Performance of TLR methods running on Rockchip RV1126 evaluated by the two test sets.}
  \vspace{-5mm}
\begin{center}
\begin{tabular}{l|lllll|lllll}
\toprule
\multirow{2}{*}{\textbf{Method}} & \multicolumn{5}{c|}{\textbf{DTLD}~\cite{8460737}} & \multicolumn{5}{c}{\textbf{NTLD}}                                        \\ 
                      & \multicolumn{1}{c|}{\textit{mAP (\%)}} & \multicolumn{1}{c|}{\textit{Precision (\%)}} & \multicolumn{1}{c|}{\textit{Recall (\%)}} & \multicolumn{1}{c|}{\textit{F1 (\%)}}& \multicolumn{1}{c|}{\textit{FPS}} & \multicolumn{1}{c|}{\textit{mAP (\%)}} & \multicolumn{1}{c|}{\textit{Precision (\%)}} & \multicolumn{1}{c|}{\textit{Recall (\%)}} & \multicolumn{1}{c|}{\textit{F1 (\%)}}& \multicolumn{1}{c}{\textit{FPS}} \\
\midrule
Two-stage: IV~\cite{7995785}                  & \multicolumn{1}{r}{53.20} & \multicolumn{1}{r}{54.19} & \multicolumn{1}{r}{53.67} & \multicolumn{1}{r}{53.93} & \multicolumn{1}{c|}{-} & \multicolumn{1}{r}{59.81}   & \multicolumn{1}{r}{59.61} & \multicolumn{1}{r} {60.33}  & \multicolumn{1}{r}{59.97} & \multicolumn{1}{r} {-}  \\  
Two-stage: ICRA~\cite{7989163}                 & \multicolumn{1}{r}{68.71} & \multicolumn{1}{r}{67.32} & \multicolumn{1}{r}{68.52} & \multicolumn{1}{r}{67.91} & \multicolumn{1}{r|}{0.71} & \multicolumn{1}{r}{73.29}   & \multicolumn{1}{r}{72.76} & \multicolumn{1}{r} {74.91}  & \multicolumn{1}{r}{73.81} & \multicolumn{1}{r} {1.14}  \\  
Two-stage: TMC~\cite{8611202}                    & \multicolumn{1}{r}{68.75} & \multicolumn{1}{r}{67.91} & \multicolumn{1}{r}{69.04} & \multicolumn{1}{r}{68.47} & \multicolumn{1}{r|}{1.35} & \multicolumn{1}{r}{74.13}   & \multicolumn{1}{r}{73.69} & \multicolumn{1}{r} {75.17}  & \multicolumn{1}{r}{74.42} & \multicolumn{1}{r} {2.37}  \\  
\midrule

One-stage: RCNN~\cite{girshick2015region}                 & \multicolumn{1}{r}{63.29} & \multicolumn{1}{r}{63.51} & \multicolumn{1}{r}{64.02} & \multicolumn{1}{r}{63.76} & \multicolumn{1}{r|}{3.16} & \multicolumn{1}{r}{68.78}   & \multicolumn{1}{r}{70.05} & \multicolumn{1}{r} {69.80}  & \multicolumn{1}{r}{69.92} & \multicolumn{1}{r} {3.75}  \\   

One-stage: YOLO~\cite{7780460}                   & \multicolumn{1}{r}{64.37} & \multicolumn{1}{r}{64.86} & \multicolumn{1}{r}{65.93} & \multicolumn{1}{r}{65.39} & \multicolumn{1}{r|}{4.83} & \multicolumn{1}{r}{71.29}   & \multicolumn{1}{r}{72.16} & \multicolumn{1}{r} {72.57}  & \multicolumn{1}{r}{72.36} & \multicolumn{1}{r} {6.94}  \\  

One-stage: SSD~\cite{liu2016ssd}               & \multicolumn{1}{r}{65.91} & \multicolumn{1}{r}{66.49} & \multicolumn{1}{r}{66.27} & \multicolumn{1}{r}{66.38} & \multicolumn{1}{r|}{5.11} & \multicolumn{1}{r}{71.67}   & \multicolumn{1}{r}{73.05} & \multicolumn{1}{r} {72.77}  & \multicolumn{1}{r}{72.91} & \multicolumn{1}{r} {6.09}  \\    

One-stage: DETR~\cite{carion2020end}                   & \multicolumn{1}{r}{70.31} & \multicolumn{1}{r}{72.09} & \multicolumn{1}{r}{71.75} & \multicolumn{1}{r}{71.92} & \multicolumn{1}{r|}{-} & \multicolumn{1}{r}{76.32}   & \multicolumn{1}{r}{76.86} & \multicolumn{1}{r} {75.49}  & \multicolumn{1}{r}{76.17} & \multicolumn{1}{r} {-}  \\   

\midrule
One-stage: ViTLR-$3$                     & \multicolumn{1}{r}{70.72} & \multicolumn{1}{r}{72.59} & \multicolumn{1}{r}{72.94} & \multicolumn{1}{r}{72.76} & \multicolumn{1}{r|}{10.55} & \multicolumn{1}{r}{79.60}   & \multicolumn{1}{r}{80.25} & \multicolumn{1}{r} {80.81}  & \multicolumn{1}{r}{80.53} & \multicolumn{1}{r} {25.16}  \\  
One-stage: ViTLR-$5$                     & \multicolumn{1}{r}{72.69} & \multicolumn{1}{r}{76.13} & \multicolumn{1}{r}{77.05} & \multicolumn{1}{r}{76.59} & \multicolumn{1}{r|}{8.50} & \multicolumn{1}{r}{85.75}   & \multicolumn{1}{r}{90.35} & \multicolumn{1}{r} {91.05}  & \multicolumn{1}{r}{90.70} & \multicolumn{1}{r} {13.21}  \\  
\bottomrule
\end{tabular}
\end{center}
  \vspace{-5mm}
\end{table*}

\section{Experiments}
\subsection{Real-world Datasets}
We adopt two real-world datasets (i.e., DTLD~\cite{8460737} and NTLD) to assess the performance of all TLR methods mentioned by this work, including the proposed {\it ViTLR} and other mainstream TLR approaches. 
DTLD (DriveU Traffic Light Dataset)~\cite{8460737} was collected from $11$ biggest cities in Germany. It contains $2110$ video clips under a frame rate of $15$Hz. More than $290,000$ traffic lights were fully annotated in each frame with an image resolution of $2048\times1024$ pixel. 
To verify the consistency of experimental results, we built another real-world dataset, NTLD (NavInfo Traffic Light Dataset), which was collected from $4$ metropolises in Mainland China. It contains $5000$ video clips under a frame rate of $15$Hz. Over $500,000$ traffic lights were fully annotated in each frame with an image resolution of $1920\times1080$ pixel. 
Reported by Table~\uppercase\expandafter{\romannumeral1}, both datasets were divided into three subsets (i.e., \textit{Train}, \textit{Valid}, and \textit{Test}) according to the ratio of $70\%$, $10\%$ and $20\%$, for the purpose of training, validating, and testing TLR approaches.

\subsection{Evaluation Metrics}
As mainstream methods on TLR generally output bounding boxes and states of traffic lights, we need to measure their locations as well as classes/categories compared with ground truths. \textit{mAP} (i.e., mean Average Precision) is a popular metric to evaluate the detection performance of the model on different traffic light classes in this work. \textit{Precision} and \textit{Recall} are leveraged to evaluate the correctness and completeness of predicted states of traffic lights, whose \textit{IoU} $>0.5$ are considered. \textit{F1} score is the harmonic mean of precision and recall, providing a single metric to balance the trade-off between the two metrics. To fulfill the requirement of real-time recognition of traffic lights for autonomous driving, \textit{FPS} (i.e., Frames Per Second) is adopted as a standard metric to measure the efficiency of TLR methods running on Rockchip RV1126.

\subsection{Comparison Results}
The mainstream approaches on TLR can be categorized
into two-stage systems (such as IV~\cite{7995785}, ICRA~\cite{7989163}, and TMC~\cite{8611202}) and one-stage methods (including SSD~\cite{liu2016ssd}, RCNN~\cite{girshick2015region}, YOLO~\cite{7780460}, and DETR~\cite{carion2020end}).
The proposed \textit{ViTLR} belongs to one-stage methods. Given that all the mainstream approaches take a single frame as inputs when conducting inference, we prepared two versions of \textit{ViTLR} for comparison. \textit{ViTLR-1} was trained by the labeled traffic lights at the current frame, and \textit{ViTLR-3} took the last $3$ consecutive frames as video-based inputs.

Table~\uppercase\expandafter{\romannumeral2} reports the evaluation results between the mainstream approaches and \textit{ViTLR}. Obviously, the proposed method has achieved the best results on both the DTLD and NTLD datasets. \textit{ViTLR-3} has achieved a mAP of $79.6\%$ while reaching an inference speed of $25$ FPS, performing better than the existing methods in recognition effect and inference efficiency.

\section{Discussions}
To further explore the pros and cons of various TLR methods, we discuss their performance from the perspectives of temporal stability, target distances, and complex scenarios in the NTLD dataset.

\subsection{Temporal Stability}
\begin{figure}[htp!]
\centering
  \includegraphics[width=0.98\columnwidth]{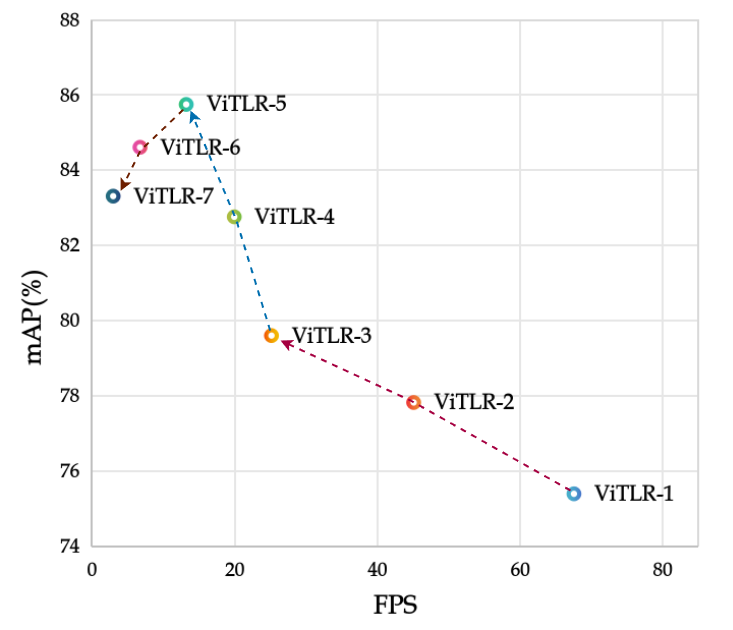}
  \vspace{-5mm}
  \caption{Performance of the TLR methods evaluated by different numbers of consecutive frames.}
  \vspace{-5mm}
  \label{fig_5}
\end{figure}

This paper verifies the stability of the \textit{ViTLR} under varying frame intervals from 1 to 7 frames, as illustrated in Figure~\ref{fig_5}. Notably, \textit{ViTLR} shows significant fluctuations in its performance at the 6-frame interval and 7-frame, indicating poor temporal stability. In contrast, \textit{ViTLR} demonstrates stable recognition effects at intervals of 2, 3, 4, and 5 frames. However, \textit{ViTLR-5} is less efficient than the \textit{ViTLR-3}. Therefore, the \textit{ViTLR-3} not only exhibits good temporal stability but also offers higher running efficiency.

\subsection{Target Distances}
\begin{figure}[htp!]
\centering
  \includegraphics[width=0.98\columnwidth]{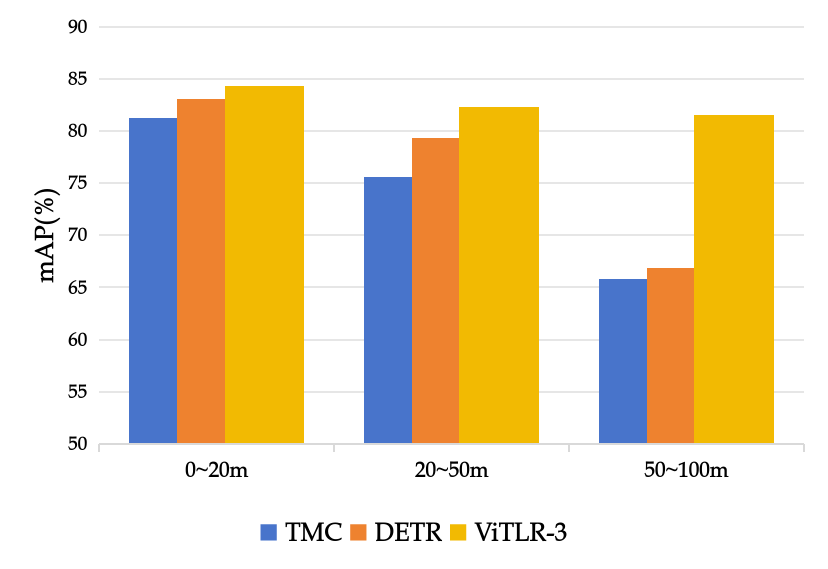}
  \vspace{-3mm}
  \caption{Performance of the TLR methods evaluated by different target distances.}
    \vspace{-5mm}
  \label{fig_6}
\end{figure}

Distance is a significant factor that influences the effectiveness of recognition. In this study, we performed a statistical analysis of the indicators for three distance intervals: less than 20 meters, between 20 and 50 meters, and between 50 and 100 meters, as illustrated in Figure~\ref{fig_6}. Our findings indicate that TMC and DETR exhibit relatively poor recognition performance for targets at greater distances, primarily due to their challenges in effectively aggregating features from different levels. In contrast, our model maintains stable performance across various distances, with no significant decrease in the indicators. 

\subsection{Complex Scenarios}
\begin{figure}[htp!]
\centering
  \includegraphics[width=0.98\columnwidth]{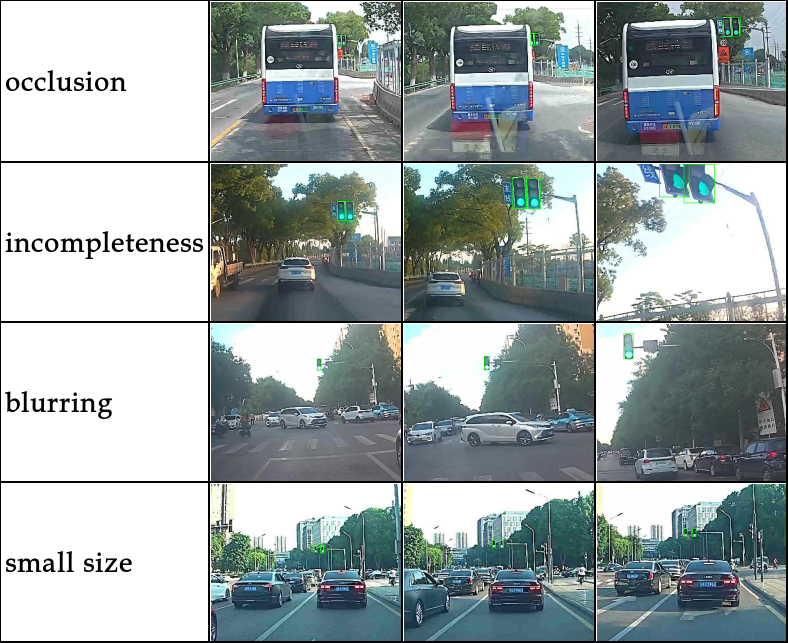}
  \caption{Performance of the TLR methods evaluated by different complex scenarios.}
    \vspace{-5mm}
  \label{fig_7}
\end{figure}

The recognition of real traffic targets faces several challenges, including occlusion, incompleteness, blurring, and small size. Figure~\ref{fig_7} provides a comparative analysis of these difficult scenarios. Our findings indicate that models utilizing multi-frame temporal features significantly outperform standard single-frame detection models, particularly in situations involving occlusion and incompleteness. The \textit{ViTLR} proposed in this paper also surpasses other models in these respects. Additionally, in the recognition of blurred and small-sized targets, the \textit{ViTLR} model demonstrates strong stability and achieves better results, even when targets appear intermittently. Consequently, our model is more suitable for locating and classifying targets across continuous frames.

\section{Conclusion}
We introduce a fundamental shift in approaching traffic light recognition by reformulating it as a video-based temporal learning problem rather than traditional single-frame analysis. This new perspective directly addresses critical limitations in existing approaches, particularly their vulnerability to temporary occlusions and lighting variations. Additionally, we tackle the challenging requirement of embedded deployment on resource-constrained hardware (Rockchip RV1126) while maintaining real-time performance - a crucial consideration often overlooked in previous studies.

To address these challenges, we propose \textit{ViTLR}, an end-to-end video-based neural architecture that processes multiple consecutive frames for robust recognition. Our solution features an embedded-optimized transformer-like architecture utilizing convolutional self-attention modules, specifically tailored for Rockchip RV1126's NPU capabilities. \textit{ViTLR} can be further enhanced through integration with HD maps, creating a complete ego-lane traffic light recognition system that bridges the gap between theoretical detection capabilities and practical autonomous driving needs.

% To validate our approach and contribute to the broader research community, we provide comprehensive evaluation resources and datasets. We introduce two substantial real-world datasets: DTLD, collected from 11 major German cities with diverse traffic light configurations, and NTLD, gathered from 4 metropolitan areas in Mainland China with different environmental conditions. Our evaluation framework goes beyond traditional metrics, analyzing temporal stability across consecutive frames, performance variations across different target distances, and robustness under complex scenarios including occlusions and backlighting. To facilitate reproducibility and future research, we make our source codes and NTLD dataset, publicly available.


%%%%%%%%%%%%%%%%%%%%%%%%%%%%%%%%%%%%%%%%%%%%%%%%%%%%%%%%%%%%%%%%%%%%%%%%%%%%%%%%

%%%%%%%%%%%%%%%%%%%%%%%%%%%%%%%%%%%%%%%%%%%%%%%%%%%%%%%%%%%%%%%%%%%%%%%%%%%%%%%%

%%%%%%%%%%%%%%%%%%%%%%%%%%%%%%%%%%%%%%%%%%%%%%%%%%%%%%%%%%%%%%%%%%%%%%%%%%%%%%%%
\section*{Acknowledgments}
This work was sponsored by the Beijing Nova Program (No. 20240484616). 
\balance
\bibliographystyle{IEEEtran}
\bibliography{IEEEexample}
\end{document}